\documentclass{article}

\usepackage{arxiv}

\usepackage[utf8]{inputenc} % allow utf-8 input
\usepackage[T1]{fontenc}    % use 8-bit T1 fonts
\usepackage{hyperref}       % hyperlinks
\usepackage{url}            % simple URL typesetting
\usepackage{booktabs}       % professional-quality tables
\usepackage{amsfonts}       % blackboard math symbols
\usepackage{nicefrac}       % compact symbols for 1/2, etc.
\usepackage{microtype}      % microtypography
\usepackage{lipsum}
\usepackage{graphicx}
\usepackage{amsmath}
\graphicspath{ {./images/} }

\title{Co-design of Neural and Muscle Network based on Embodied Perceptron Representation
\thanks{
Published in the Proceedings of the IEEE/SICE International Symposium on System Integration (SII 2026).\\
DOI: 10.1109/SII64115.2026.11404727.\\
\textcopyright{} 2026 IEEE. Personal use of this material is permitted. Other reuse requires permission from IEEE; see the IEEE posting policy for details.}
}

\author{
 Siyuan Tao \\
  Department of Mechanical Engineering\\
  The University of Osaka, Osaka \\
  \texttt{suen.tou@eom.mech.eng.osaka-u.ac.jp} \\
   \And
 Yoichi Masuda \\
  Department of Mechanical Engineering\\
  The University of Osaka, Osaka \\
  \And
 Hiroyuki Nabae \\
  Department of Mechanical Engineering\\
  Institute of Science Tokyo, Tokyo \\
  \And
 Masato Ishikawa \\
  Department of Mechanical Engineering\\
  The University of Osaka, Osaka \\
}

\begin{document}
\maketitle
\begin{abstract}
Recent advances in AI technologies have enabled the advanced design of complex control policies. In contrast, focusing on the body, many robots still employ simple bodies that can limit adaptability to environments.
Studies in embodied robotics have shown that well-designed bodies can partially replace the role of control and computation with physical body--environment interactions, yet such designs still depend heavily on expert intuition. 
% Key challenges include the absence of a systematic and theoretical framework for body design and a method for jointly optimizing the body and controller.
There is a need for a systematic theoretical framework for body design, as well as a method for joint optimization of the body and controller.
To address this, we introduce the Embodied Perceptron, a theoretical framework that unifies neural networks and physical body systems. 
In this view, the body itself acts as a perceptron: mechanical parameters correspond to weights, and physical nonlinearities play the role of activation functions.
By representing physical constraints as weights and nonlinear properties as activation functions, a physical body can be modeled in neural-network form.
The system representation enables us to explicitly and theoretically explain that the body can substitute for part of the neural control.
% This system representation provides a foundation for the design and evaluation of embodied intelligence, enabling the co-design of physical body and controller within dynamic tasks.
As an application, we co-optimize control policy and muscle configuration in a musculoskeletal robot and show that the resulting embodied intelligence can provide inherent stability, improve learning efficiency, and drastically reduce model size—even with a single-neuron controller.
% This system representation, which bridges the informational and physical worlds, explicitly and theoretically explains that the body can substitute for part of the neural control, providing a pathway toward understanding and systematically designing embodied AI systems.
The results bridge the informational and physical worlds and provide a pathway toward understanding and systematic design of embodied AI systems.
\end{abstract}

% keywords can be removed

\section{Introduction}
Recent advances in AI technologies, particularly neural networks, have led to advanced methods for designing complex, high-performance control systems. In contrast, when we turn our attention to robot bodies, many existing robots still follow the “simplicity is best” philosophy, adopting traditional serial-link structures built from motors and rigid links.
While such simple bodies facilitate design, they may limit a robot’s ability to adapt to complex environments.
In contrast, studies in embodied robotics\cite{ref1} and bio-inspired robotics\cite{ref2} have shown that appropriately designed bodies can simplify control and enhance adaptability to environments, by partially replacing the role of control and computation with physical body--environment interactions. However, designing such bodies still depends heavily on the designer’s expertise, creating challenges for both design and control.

There are several challenges in designing such advanced embodied systems. First, the functional role of the body itself—in terms of control and information processing—remains unclear. To better understand the computational capabilities that arise from body–environment interactions, researchers have explored physical reservoir computing. In this approach, linear regression applied to the output of a physical body enables the learning of nonlinear filters\cite{ref3}. Other studies have used control theory to clarify how body structure relates to control functions, such as in the self-stabilization of passive dynamic walkers\cite{ref4}.
Second, there is the challenge posed by the lack of a unified, theory-driven design framework that gives equal consideration to the controller and the body, enabling their joint optimization through movement. In studies along this line, researchers have investigated tendon-driven underactuated robots using reinforcement learning\cite{ref5}, and the determination of optimal multi-articular muscle arrangements for specific robot motions based on human motion data\cite{ref6}. 
One approach for treating the controller and the body on an equal footing is to model hardware as a policy within a reinforcement learning framework. By representing the body as an auto-differentiable computational graph, both mechanical and computational parameters can be optimized simultaneously through gradient-based methods\cite{ref7}. However, current representations remain limited to specific computational graph formulations.

In this paper, we introduce the \textit{Embodied Perceptron representation}, a theoretical framework for describing general neural networks and physical body systems in a unified manner. 
By modeling physical constraints as weights and nonlinear physical properties as activation functions, physical body systems can be represented in the form of neural networks.
As an application of this system representation, we present the co-optimization of control and muscle configuration in a musculoskeletal robot. The musculoskeletal system, with its nonlinear muscle dynamics, plays a crucial role in providing inherent stability and advantageous properties for learning\cite{ref8}. In this study, we demonstrate that a well-designed musculoskeletal system can significantly reduce the size of the controller.
This approach aims to bridge the informational world and the physical world, providing a pathway toward understanding and the systematic design of embodied AI systems.

\section{THEORY}
The Embodied Perceptron provides a unified representation of physical and neural systems. Using this representation, several simple physical systems can be described as neural networks\cite{ref9}.
The perceptron analogy is not only metaphorical but provides a formal link between mechanics and control.

\subsection{Forward Propagation in Neural Networks}

Consider a fully connected feedforward neural network with $L$ layers. Let $\mathrm{W}^{(l)} \in \mathbb{R}^{n_l \times n_{l-1}}\; (l=1,...,L)$ denote the weight matrix connecting layer $l-1$ to layer $l$, where $n_l$ is the number of neurons in layer $l$. Let $\mathit{b}^{(l)} \in \mathbb{R}^{n_l}$ be the bias vector for layer $l$, and $\mathit{h}^{(l)} \in \mathbb{R}^{n_l}$ represent the output activations of layer $l$. The network input is denoted as $\mathit{h}^{(0)} \in \mathbb{R}^{n_0}$.
Then the forward propagation through layer $l$ is computed as:
\begin{align}
\mathit{z}^{(l)} &= \mathrm{W}^{(l)}\mathit{h}^{(l-1)} + \mathit{b}^{(l)}, \label{eq:linear_transform}\\
\mathit{h}^{(l)} &= \phi\left(\mathit{z}^{(l)}\right), \label{eq:activation}
\end{align}
where $\mathit{z}^{(l)} \in \mathbb{R}^{n_l}$ represent the pre-activation values, and $\phi(\cdot): \mathbb{R} \to \mathbb{R}$ is an element-wise non-linear activation function.

\subsection{Analogy between the Musculoskeletal Systems and Neural Networks}

Consider a musculoskeletal system with $n$ degrees of freedom joints and $m$ muscles. For simplicity, assume that muscles are connected to pulleys attached to joints.
In this musculoskeletal system, the length of each muscle $\boldsymbol{\ell} \in \mathbb{R}^m$ is written as follows using joint angles $\boldsymbol{\theta} \in \mathbb{R}^n$:
\begin{align}
\boldsymbol{\ell} - \boldsymbol{\ell}(0) \approx \left. \frac{\partial \ell}{\partial \theta} \right|_{\theta(0)} (\theta - \theta(0)) = \mathrm{R}(\boldsymbol{\theta} - \boldsymbol{\theta}(0)),
\end{align}
where $\boldsymbol{\ell}(0)$ is the initial muscle length at initial joint angles $\boldsymbol{\theta}(0)$, and $\mathrm{R} \in \mathbb{R}^{m \times n}$ is a Jacobian matrix. 
The positions of non-zero elements in the matrix indicate the connections between muscles and joints, and their values represent the lengths of the moment arms based on pulley radii.
With initial joint angles $\boldsymbol{\theta}(0) = \mathbf{0}$, the muscle extension $\Delta\boldsymbol{\ell} = \boldsymbol{\ell} - \boldsymbol{\ell}_0$ from the natural length $\boldsymbol{\ell}_0 \in \mathbb{R}^m$ can be written as the following neural network forward propagation form:
\begin{align}
\mathit{z}^{(1)} = \mathrm{R}\theta + \mathit{b} = \Delta \ell,
\end{align}
where $\mathit{b} = \ell(0) - \ell_0$ is the bias term. For example, when learning is applied to Eq. (4), the natural length can be obtained by calculating $\ell_0 = \ell(0) - \mathit{b}$.

Next, we describe the relationship between muscle force and torque as a neural network. The tension $\mathit{f} \in \mathbb{R}^m$ generated by each muscle acts on the joints as torque $\tau \in \mathbb{R}^n$:
\begin{align}
\tau = \mathrm{R}^{\top}\mathit{f}.
\end{align}

Here, we consider the case where muscle tension $f$ arises from muscle force $\mathit{f}_{\text{passive}}$ generated by passive elements of the muscle model and muscle force $\mathit{f}_{\text{active}}$ generated by active elements. 
We also assume that the muscle model does not generate force during shortening, and that the passive element behaves as a first-order nonlinear spring. In this case, the joint torque $\tau$ can be described as follows:
\begin{align}
\tau &= \mathrm{R}^{\top}(\mathit{f}_{\text{passive}} + \mathit{f}_{\text{active}}) \\
&= \mathrm{R}^{\top}(-\phi(\mathrm{K}\Delta \ell) + \mathit{f}_{\text{active}}),
\end{align}
where $\phi(\cdot)$ is a variant of the ReLU function, and $\mathrm{K} \in \mathbb{R}^{m \times m}$ is a diagonal stiffness matrix, which contains positive stiffness coefficients.

% From Eqs. (4) and (7), the musculoskeletal system with joint angles as input and joint torques as output can be written as the following neural network forward propagation form:
From Eqs. (4) and (7), the musculoskeletal system can be modeled as a neural network mapping joint angles to torques:
\begin{align}
\mathit{z}^{(1)} &= \mathrm{R}\boldsymbol{\theta} + b \\
\mathit{z}^{(2)} &= -\phi(\mathrm{K}z^{(1)}) + f_{\text{active}} \\
\mathit{z}^{(3)} &= \mathrm{R}^{\top}z^{(2)}\\
\tau &= z^{(3)}.
\end{align}
Here, the active muscle force $\mathit{f}_{\text{active}}$ is the output from another neural network corresponding to the control policy.

It is important that the unilateral nature of muscles inherently acts as nonlinear activation functions, fundamentally integrating physical constraints into computation. 
Recognizing this mechanism deepens our theoretical understanding of biorobotic symbiosis, which refers to the reciprocal relationship between biological systems and robotic design.

\subsection{Representing Nonlinear Muscle Properties via Activation Functions}

% In this study, we investigated the role of activation functions in shaping the nonlinear muscle properties. Two types of functions were tested: the Rectified Linear Unit (ReLU) and its squared variant, ReLU\textsuperscript{2}.
In this study, nonlinear muscle properties are modeled as activation functions. Among various candidates, we adopt ReLU\textsuperscript{2}, a squared variant of ReLU(Rectified Linear Unit).

The ReLU function, defined as $\mathrm{ReLU}(x) = \max(0, x)$, is widely used in deep learning for its computational simplicity and ability to mitigate vanishing gradients. In this work, we use a variant of ReLU to approximate muscle behavior with slack, where the muscle generates no force below a certain extension.
Specifically, the ReLU\textsuperscript{2} function, defined as $\mathrm{ReLU}^2(x) = \max(0, x)^2$, is less common in deep learning but has been used to model biological muscle properties, including nonlinear stiffness and the force–length relationship of passive tissues\cite{ref10}. Here, we adopt it as the activation function for passive force.

% In the context of this study, the ReLU\textsuperscript{2} function was employed to model the nonlinear elastic behavior of muscles, inspired by biological observations.

\section{METHODS}

\subsection{Experimental Setup and Procedure}

\setcounter{figure}{0} 
\begin{figure}[t]
    \centering
    \includegraphics[scale=0.26]{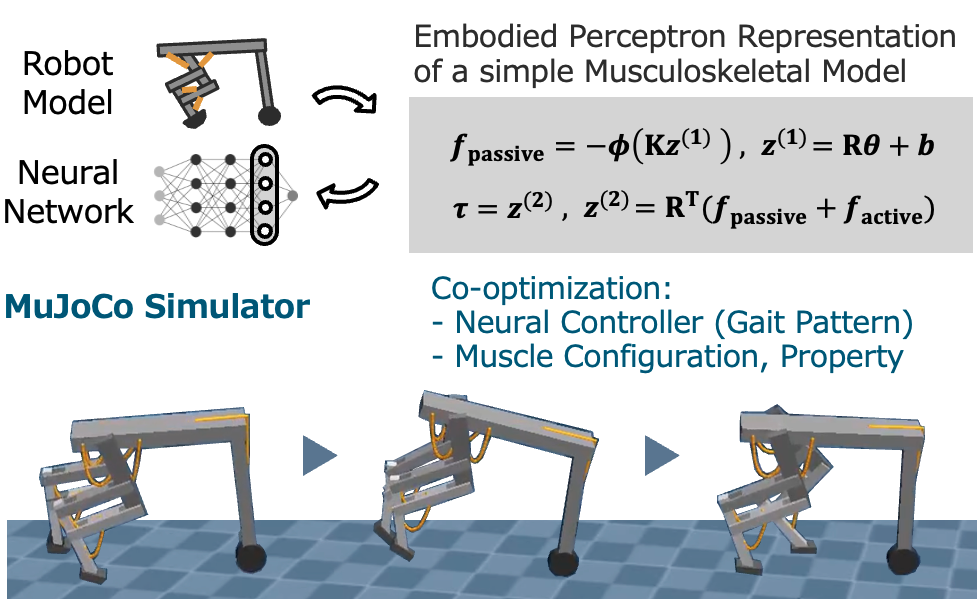}
    \caption{Experimental procedure.}
    \label{fig1}
    \vspace{-1em}
\end{figure}

\setcounter{figure}{1} 
\begin{figure}[t]
    \centering
    \includegraphics[scale=0.3]{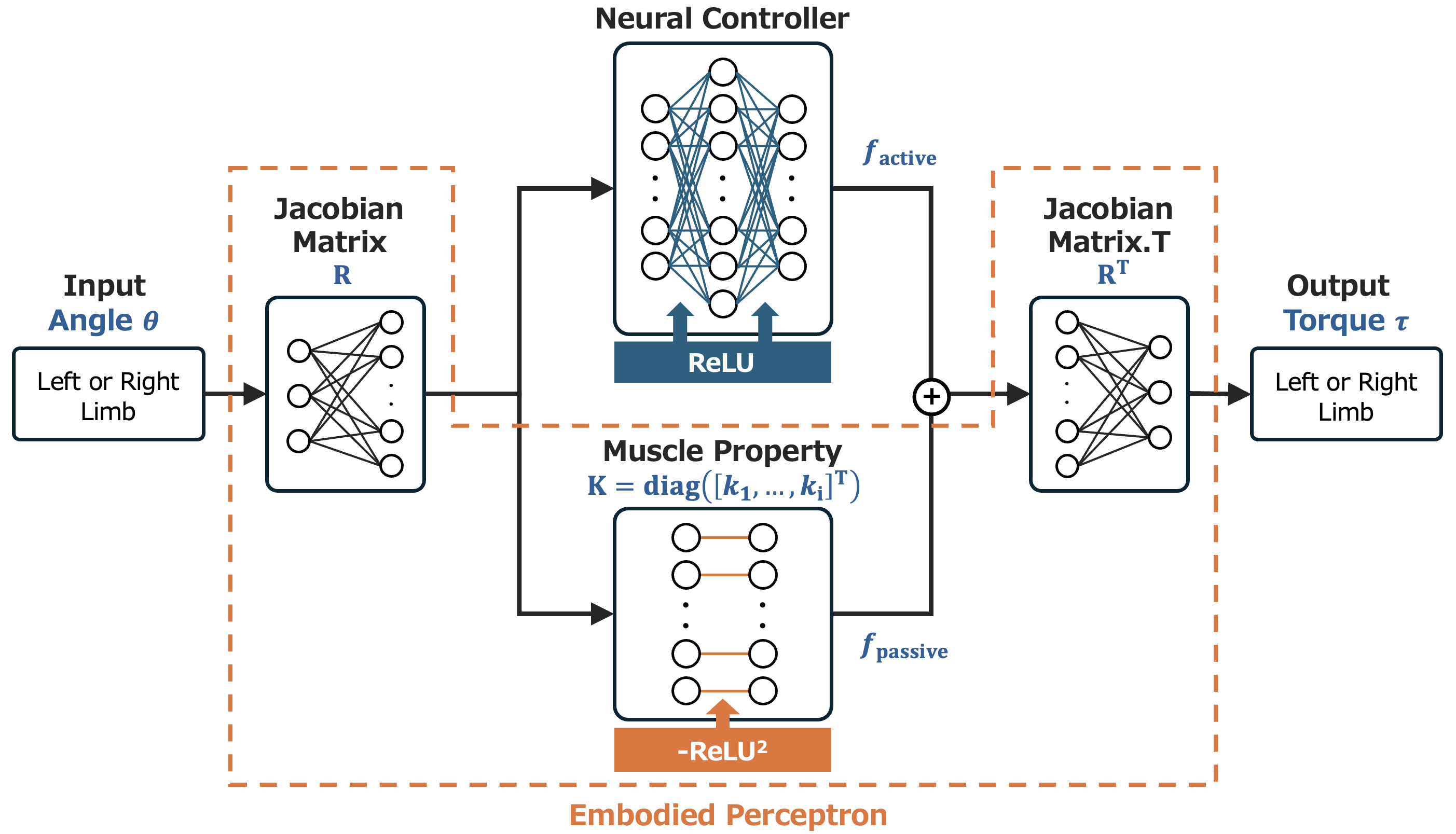}
    \caption{Overview of the system represented by the Embodied Perceptron. The orange dashed region indicates the physical muscle network, while the neural controller at the top represents the control policy network. In the orange dashed region, $\mathrm{R}$ denotes the muscle configuration, and $\mathrm{K}$ captures the muscle properties, including the nonlinear elasticity of each muscle.} 
    \label{fig2}
    \vspace{-.7em}
\end{figure}

In this study, we proposed the Embodied Perceptron representation that integrates the control policy and physical structure of a simplified musculoskeletal robot model.
This model was implemented in the MuJoCo\cite{ref11} and jointly optimized three aspects: the neural control policy, the muscle configuration and muscle properties, as depicted in Fig.~\ref{fig1}.

Figure \ref{fig2} shows the integrated body--control system represented by the Embodied Perceptron. 
% Specifically, the robot is modeled as a neural network that maps motor commands to muscle forces, which are then transformed into joint torques through linear mappings and activation functions.
The model is a neural network that maps motor commands to muscle forces, which are then transformed into joint torques through linear mappings and activation functions. The orange dashed area represents the physical muscle network, and the network at the top of the figure represents the control policy. 
In the orange dashed area, the Jacobian matrix $\mathrm{R}$ corresponds to muscle configuration, and the matrix $\mathrm{K}$ at the bottom captures muscle properties. 
By training $\mathrm{R}$ and $\mathrm{K}$, collectively referred to as the Embodied Perceptron in this study, the network learns a body structure adapted to walking. 
Simultaneously, optimizing the layers of the neural controller enables the acquisition of an effective locomotion policy.

Furthermore, we employed the CMA-ES\cite{ref12} as the optimization algorithm to search for optimal parameters. 
Unlike gradient-based methods, CMA-ES does not rely on differentiable objectives and is more robust to noisy or discontinuous reward landscapes, which are often encountered in co-optimization of morphology and control.
All experiments were conducted on an Ubuntu 22.04 workstation with an Intel Core i9-14900K processor.
%During initialization, joint angles were randomly sampled within the ranges (in radians) specified in the Fig.~\ref{fig1_5} to encourage the emergence of diverse gait patterns.
During initialization, joint angles were randomly sampled within the ranges (image posture as zero angles; radians; extension positive, flexion negative) specified in Fig.~\ref{fig1_5}.
At the same time, the random seed was fixed to ensure reproducibility and enable fair comparison of the experimental results in various conditions.

\subsection{Reward Function}

In this study, we designed a reward function that motivates the robot to maximize its walking distance within a fixed time horizon while minimizing energy consumption and penalizing undesirable weight configurations. 
Given:
\begin{itemize}
    \item $T$: total number of time steps,
    \item $q_t$: walking distance at time step $t$,
    \item $\tau_t$: joint torques at time step $t$ (6-dimensional),
    \item $v_t$: joint velocities at time step $t$,
    \item $\mathrm{R}$: weights in the neural controller,
    \item $\mathrm{K}$: muscle properties,
    \item $\alpha$: hyperparameter for energy penalty,
    \item $\beta_1$, $\beta_2$: hyperparameters for negative weight penalties,
    \item $\gamma$: hyperparameter for L1 regularization.
\end{itemize}
\textbf{(1) Energy Consumption Penalty:}
To discourage inefficient energy usage, the following energy penalty is applied:
\begin{align}
P_\mathrm{E} = \frac{\alpha}{T} \sum_{t=1}^T \sum_{i=1}^6 \left| \tau_{t,i} \cdot v_{t,i} \right|.
\end{align}
\textbf{(2) Negative Weight Penalty:}
To encourage non-negative weight configurations in the network, penalties are imposed on negative elements of $\mathrm{R}$ and $\mathrm{K}$:
\begin{align}
P_{\text{neg}} = \beta_1 \sum_{\mathrm{R}_{ij} < 0} |\mathrm{R}_{ij}| + \beta_2 \sum_{\mathrm{K}_{kl} < 0} |\mathrm{K}_{kl}|.
\end{align}
\textbf{(3) L1 Regularization:}
To promote sparsity and control the positive weights in $\mathrm{R}$, we apply an L1 penalty:
\begin{align}
P_{\text{L1}} = \gamma \sum_{\mathrm{R}_{ij} > 0} \mathrm{R}_{ij}.
\end{align}
\textbf{(4) Final Reward Function:}
Using Eqs. (12)-(14) The final reward function is defined as:
\begin{align}
r = q_T - \left(P_\mathrm{E} + P_{\text{neg}} + P_{\text{L1}}\right),
\end{align}
where $q_T$ denotes the total walking distance achieved by the robot at the final time step $T$.
The success criterion for the walking task in this study is defined as achieving a walking distance exceeding 15m within 5000 simulation steps.

\subsection{Robot Design}

In our design, we used a front-wheeled, rear-legged hybrid robot shown in Fig.~\ref{fig1_5}.
The robot was equipped with ten muscles per leg, leading to a body configuration matrix $\mathrm{R}$ of size 10$\times$3 and a diagonal matrix $\mathrm{K}$ of size 10$\times$10, as illustrated in Fig.~\ref{fig2}.
This hybrid morphology was chosen to investigate how wheeled and legged components can be integrated within a neuromechanical modeling framework to achieve adaptive locomotion.
In this work, the muscle configuration was represented by the Jacobian matrix $\mathrm{R}$, where a positive element at row $i$ and column $j$ indicates that muscle $i$ acts as an extensor for joint $j$.
\begin{align}
    \mathrm{R} = \begin{bmatrix}
    r_{0} & 0 & 0 \\
    0 & r_{1} & 0 \\
    0 & 0 & r_{2} \\
    r_{3} & r_{4} & 0 \\
    -r_{5} & -r_{6} & 0 \\
    0 & r_{7} & r_{8} \\
    0 & -r_{9} & -r_{10} \\
    -r_{11} & 0 & 0 \\
    0 & -r_{12} & 0 \\
    0 & 0 & -r_{13}
    \end{bmatrix}\;\;\; \left(r_{i}>0\right).
\end{align}

It should be noted that since no upper bound is imposed on the magnitude of each element in $\mathrm{R}$, representing the pulley radius, solutions with excessively large pulley radii may arise.
In such cases, it is necessary to apply a scaling using an equivalence transformation\footnote{
From Eqs. (4), (7), and (17),
\begin{align}
\tau &= \mathrm{R}^{\top}\Big( -\mathrm{ReLU}^2 \big( \mathrm{K}(\mathrm{R}\theta + \mathit{b})\big) + \mathit{f}_{\text{active}}\Big)\\
 &= a\mathrm{R}^{\top}\Big( -\frac{1}{a}\mathrm{ReLU}^2 \big( \frac{\mathrm{K}}{a}(a\mathrm{R}\theta + a\mathit{b})\big) + \frac{\mathit{f}_{\text{active}}}{a}\Big)\\
 &= a\mathrm{R}^{\top}\Big( -\mathrm{ReLU}^2 \big( a^{-\frac{3}{2}}\mathrm{K}(a\mathrm{R}\theta + a\mathit{b})\big) + a^{-\frac{3}{2}}\mathit{f}_{\text{active}}\Big),
\end{align}
where we use the relationship $\frac{1}{a}\mathrm{ReLU}^2(\frac{x}{a}) = \frac{1}{a}\max(0, \frac{x}{a})^2 = \max(0, \frac{x}{a^{\frac{3}{2}}})^2 = \mathrm{ReLU}^2(a^{-\frac{3}{2}}x)$.
}.
Equation (7) can be equivalently transformed by choosing a suitable positive scalar $a$ as below.
\begin{align}
    \hat{\mathrm{R}}=a\mathrm{R},~~\hat{\mathrm{K}}=a^{-\frac{3}{2}}\mathrm{K},~~\hat{b}=ab,~~\hat{f}_{\text{active}}=a^{-\frac{3}{2}}f_{\text{active}}.
\end{align}

For example, if all pulley radii are desired to be kept below $0.1$ but the maximum element of $\mathrm{R}$ after optimization is $r_{\max}>0.1$, then by choosing $a = 0.1/r_{\mathrm{max}}$, the pulley radii are scaled so that their maximum value becomes $0.1$.
%For example, if the maximum value of the elements in $\mathrm{R}$ (pulley radius) is $r_{\max}$, and all pulley radii are desired to be kept below $0.1$, then selecting
%\begin{align}
%    a=\dfrac{0.1}{r_{\mathrm{max}}}.
%\end{align}
%scales the pulley radii so that their maximum value is 0.1.
\setcounter{figure}{2} 
\begin{figure}[t]
    \centering
    \includegraphics[scale=0.21]{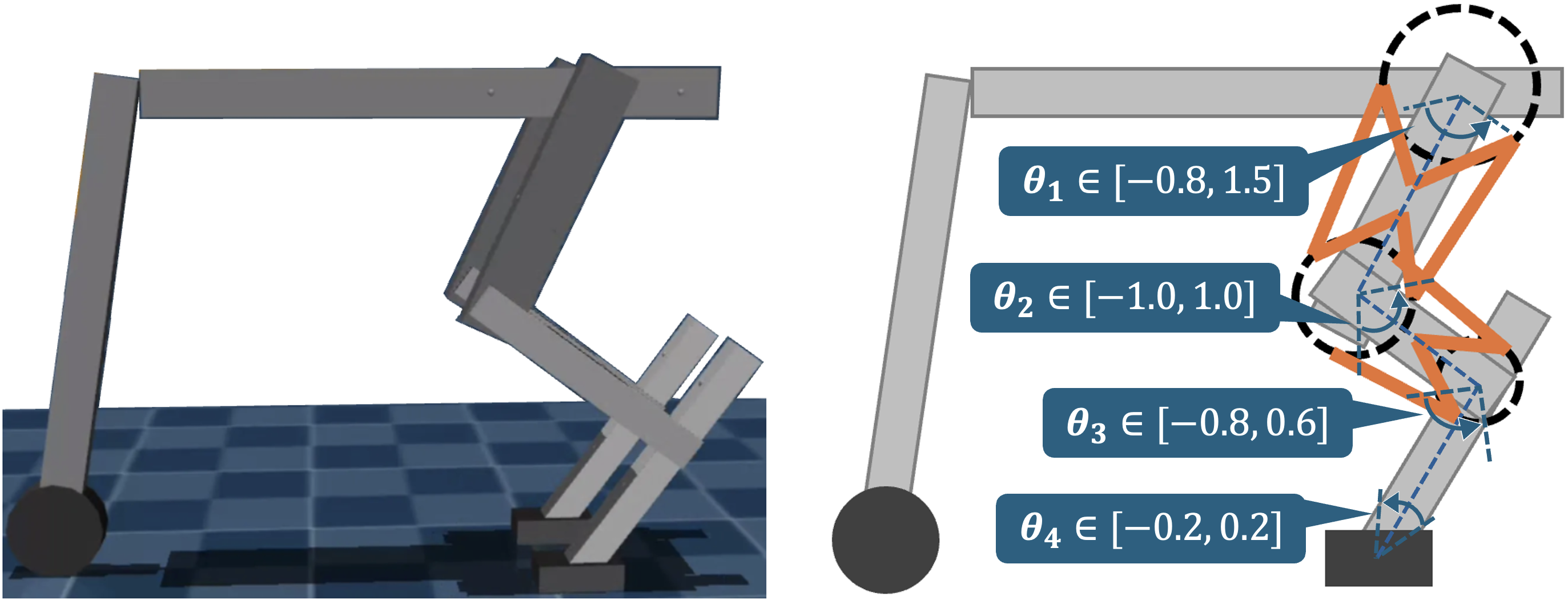}
    \caption{Robot design in this study. Left: 3D model in MuJoCo simulation showing the corresponding physical implementation of the design. Right: Schematic illustration of the leg mechanism, highlighting the joint configuration and muscle-inspired actuator layout (orange lines).}
    \label{fig1_5}
    \vspace{-.5em}
\end{figure}

\section{RESULTS}

\subsection{Impact of the number of neurons in the neural controller on gait pattern}

\begin{figure}[!b]
  \centering
  \includegraphics[scale=0.35]{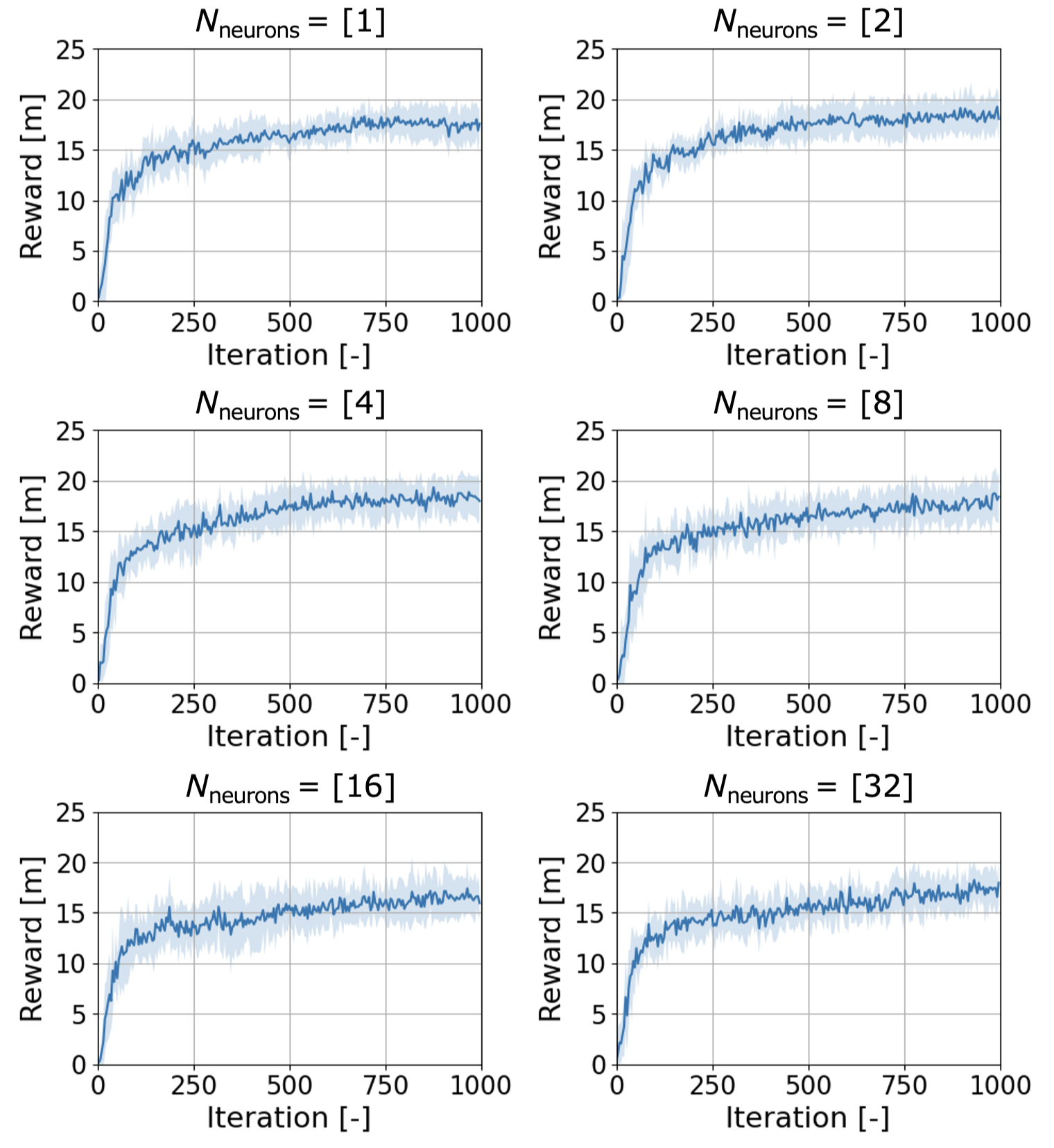}
  \caption{Impact of neural controller size on gait pattern. Network structure denoted as $N_\mathrm{neurons} =[n_1 ,n_2 ,…]$ indicates the number of neurons per layer.}
  \label{fig3}
\end{figure}

In this experiment, we varied the number of neurons in the neural controller to investigate its effect on gait patterns.
Figure~\ref{fig3} presents the training curves for hidden layer sizes ranging from 1 to 32 neurons over 1000 training iterations.
Each subplot illustrates the network’s learning progress in walking distance, averaged over ten runs, with error bars indicating the variability across trials.
Regardless of the size of the network, the reward values generally increase and stabilize with time, demonstrating the effectiveness of the proposed framework in learning the walking task.

Remarkably, the results indicate that the robot can successfully learn to walk even with extremely small networks.
As shown in the plots, networks with as few as one or two neurons in the hidden layer are capable of achieving stable learning, and long travel distance over a fixed number of steps. Additionally, reward levels comparable to those of larger networks.
This outcome underscores the role of physical constraints embedded in the robot’s body design.
Examination of the travel distance at 1000 iterations reveals that increasing the number of neurons (greater than 16) tends to slightly reduce learning efficiency.
Overall, the findings highlight the ability of the proposed architecture to achieve efficient locomotion using minimal neural resources.

\subsection{Robot structure designed through co-opimization}
\begin{figure}[b]
    \centering
    \begin{align}
    \mathrm{R} = \begin{bmatrix}
        0.014 & 0.000 & 0.000 \\
        0.000 & 0.086 & 0.000 \\
        0.000 & 0.000 & 0.052 \\
        0.031 & 0.110 & 0.000 \\
        -0.099 & -0.083 & 0.000 \\
        0.000 & 0.053 & 0.030 \\
        0.000 & -0.003 & -0.055 \\
        -0.125 & 0.000 & 0.000 \\
        0.000 & -0.121 & 0.000 \\
        0.000 & 0.000 & -0.053 \\
        \end{bmatrix},
    \mathrm{K} = \mathrm{diag}{\left(
        \begin{bmatrix}
        323.13 \\
        617.18 \\
        296.07 \\
        94.18 \\
        1391.27 \\
        4391.81 \\
        4617.86 \\
        3807.32 \\
        5724.71 \\
        7161.96 \\
        \end{bmatrix}\right)}
    \end{align}.
\end{figure}

\begin{figure}[!b]
  \centering
  \includegraphics[scale=0.4]{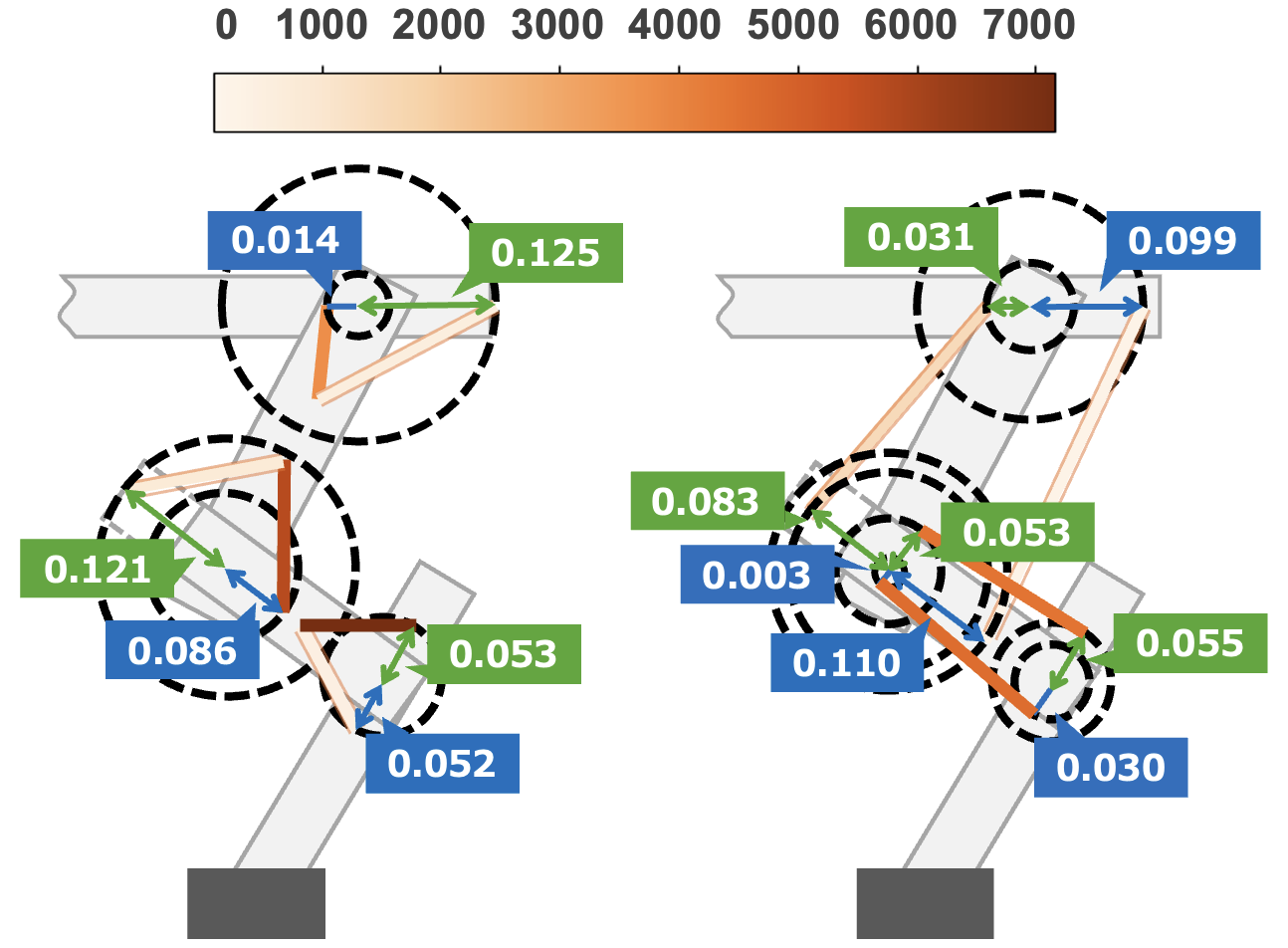}
  \caption{Optimized robot design obtained using the proposed method. Left: mono-articular muscle configuration; right: bi-articular muscle configuration. Muscles are shown in orange, with color intensity corresponding to their elasticity values (darker colors indicate higher elasticity).}
  \label{fig6}
\end{figure}

This section illustrates the optimized body structure obtained through the proposed co-design method. 
In particular, we focus on the case with a single neuron in the neural controller and present the corresponding muscle configuration Jacobian matrix $\mathrm{R}$ and muscle property matrix $\mathrm{K}=\mathrm{diag}([k_1, ..., k_{10}]^\top)$, as shown in Eq.~(21). 
One of the optimized musculoskeletal structures in Experiment A is visualized in Fig.~\ref{fig6}, where the left panel depicts the optimized configuration of mono-articular muscles, and the right panel shows the optimized configuration of bi-articular muscles.
In the figure, muscles are shown in orange, with color intensity representing their elasticity values.
These results demonstrate the effectiveness of the proposed approach in optimizing the physical body and neural controller design.

\subsection{Comparison of learning performance across neural network architectures with and without body design}

\begin{figure}[b]
  \centering
  \includegraphics[scale=0.33]{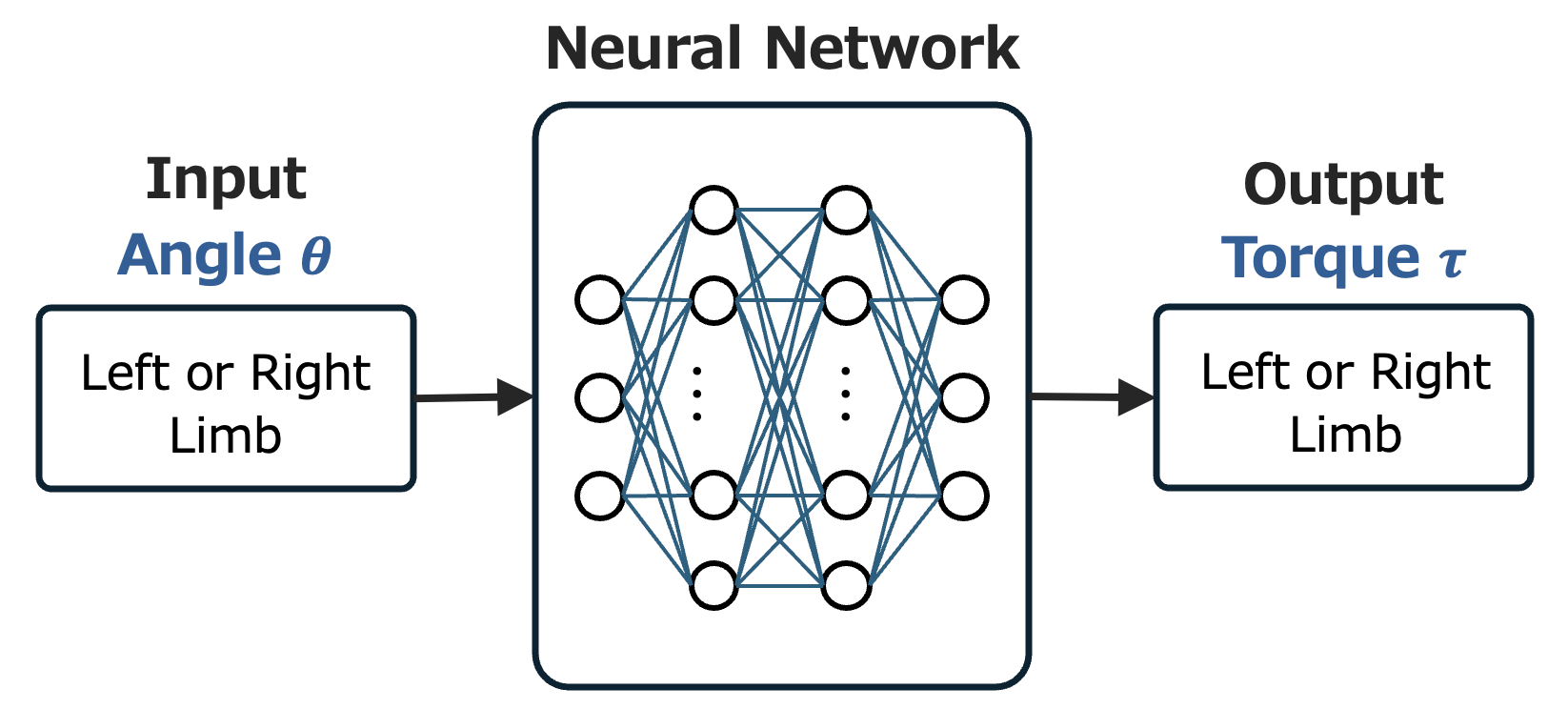}
  \caption{Conventional two-layer feedforward neural networks.}
  \label{fig4}
\end{figure}
\begin{figure}[!b]
  \centering
  \includegraphics[scale=0.38]{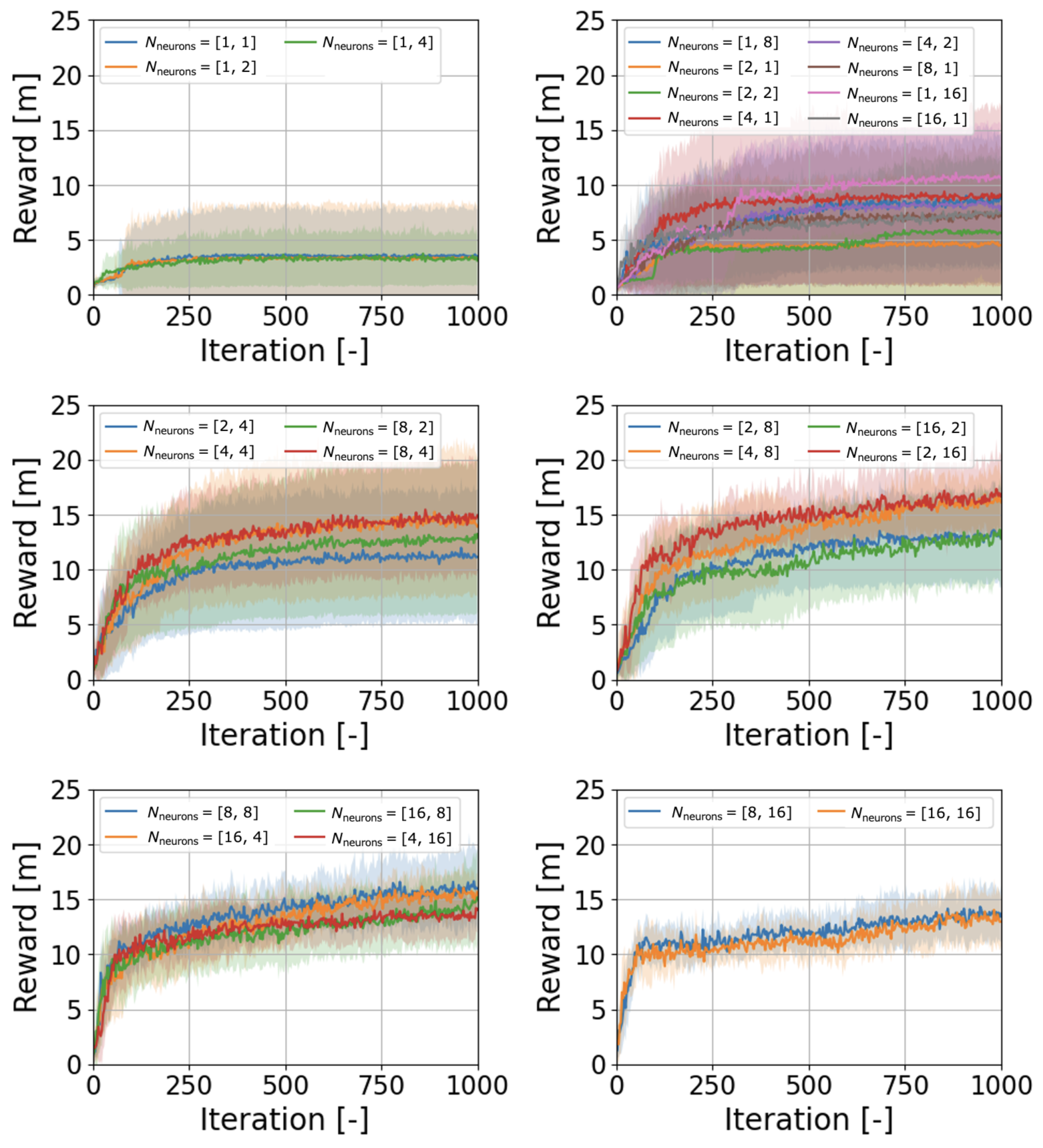}
  \caption{Impact of neural controller size in the controller-only network on gait pattern. Network structure denoted as $N_\mathrm{neurons} =[n_1 ,n_2 ,…]$ shows the number of neurons per layer.}
  \label{fig5}
\end{figure}

In this experiment, we evaluated the efficiency of the proposed Embodied Perceptron architecture by comparing it with conventional two-layer feedforward neural networks in a walking task, as shown in Fig.~\ref{fig4}.
To establish a baseline, we conducted a systematic evaluation of standard two-layer neural networks by varying the number of neurons in each layer, as illustrated in Fig.~\ref{fig5}. 
The results show that these conventional architectures required significantly more neurons, often exceeding 16 in total, to achieve comparable performance. In contrast, smaller networks exhibited poor learning efficiency and limited convergence.
On the other hand, as shown in Fig.~\ref{fig3}, the proposed architecture achieved stable learning of locomotion behaviors with as few as 1–4 neurons, demonstrating the expressive power and effectiveness of the body-aware controller design.
These findings highlight that incorporating physical body design substantially reduces the representational burden on the neural controller, enabling robust performance with significantly fewer parameters. This suggests a promising pathway toward memory-efficient and data-efficient learning through embodied co-design.

\section{DISCUSSION}
The proposed Embodied Perceptron representation enabled the explicit modeling of the body functions as components of neural networks. The model shows that the physical body is not merely as a passive structure but a computational resource supporting control.

The experimental results based on the proposed system representation offer several important insights into the role of embodied intelligence as a computational resource.
First, Experiment A examined how the number of neurons in the neural controller affects learning efficiency.
The results demonstrate that the robot whose muscle configuration and properties were jointly optimized can achieve efficient locomotion with only a single neuron in the neural controller.
Note that, as shown in Experiment B, the network described as an embodied perceptron can be realized as a physical mechanism of the robot, with the robot's computer required to execute only the neural controller.
% Additionally, the observation that a larger number of neurons can in fact reduce locomotion efficiency resembles the network characteristics of the spinal control in neurophysiology, which operates with only a small number of motor neurons.

Second, Experiment C compared learning performance across network architectures with and without body-aware design, highlighting the advantage of the Embodied Perceptron over conventional feedforward networks. While standard two-layer networks required as many as sixteen neurons to reach comparable locomotion performance, the embodied architecture achieved efficient walking with just a single neuron. This substantial reduction in neural resource demands demonstrates how incorporating body-aware design alleviates the representational burden on the controller, promoting memory-efficient and data-efficient learning.

Together, Experiments A and C reveal that effective walking behavior can be realized with remarkably minimal neural architectures. The robot successfully learned to walk using as few as one neuron in the controller, highlighting the efficiency gained by embedding physical constraints within the body design. This minimalism parallels biological systems, where simple neural circuits, such as spinal reflexes in animals, generate complex gait patterns through body dynamics. This may also suggest that inter-joint coordination naturally emerges from the body--environment interaction without requiring complex neural control[13, 14].
% \cite{ref13, ref14}.

%Overall, the results support the hypothesis that intelligence emerges from integrating neural controllers, morphologies, and the environment.
Overall, the proposed framework explicitly and theoretically shows that the body can substitute for part of the neural control, and that the resulting embodied intelligence can reduce model size and the computational load.
These findings also highlight the importance of appropriately co-designing neural and physical structures in an integrated manner.
% Overall, the results support the hypothesis that an appropriately designed body can reduce model size and the robot's computational load by substituting for part of the neural control.
% These findings highlight the effectiveness of our framework for analyzing embodied intelligence and underscore the value of co-designing neural and physical structures to develop efficient and robust robotic systems.

\section{CONCLUSION}

This paper introduced the Embodied Perceptron representation, a unified system representation of robot's neural controller and physical body. In this framework, the body itself is explicitly represented as a computational resource.
The framework also provides a foundation for co-design of the controller and robot's body.
The experiments demonstrate that effective locomotion can emerge from minimal architectures with a single-neuron controller, highlighting how embodied intelligence arises from the interaction between simple neural structures and body–environment interaction.
%The result shows that the model enables stable locomotion learning and 

Although this study focused on a specific locomotion task, the framework is general and can be extended to goal-directed motions by redefining the reward and control objectives. 
Future work will scale it to more complex robots, integrating diverse physical constraints and environmental factors to advance embodied intelligence and autonomous robotic design.
Currently, the penalty-based body parameter optimization uses soft constraints, which can occasionally yield infeasible configurations. While increasing optimization iterations can mitigate this, selecting optimal hyperparameters remains a key challenge.

\section*{ACKNOWLEDGMENT}

%%%%%%%%%%%%%%%%%%%%%%%%%%%%%%%%%%%%%%%%%%%%%%%%%%%%%%%%%%%%%%%%%%%%%%%%%%%%%%%%

This research was supported by JST FOREST Grant Number JPMJFR202F, JSPS KAKENHI 23H05445, 24H00294, and 23K22700.

\bibliographystyle{unsrt}  


\end{document}